\PassOptionsToPackage{numbers,sort&compress}{natbib}
\RequirePackage{latexml}
\iflatexml
  \documentclass{article}
  \usepackage{amsmath,amssymb,graphicx,xcolor}
  \usepackage[numbers,sort&compress]{natbib}
  \usepackage{hyperref}
\else
  \documentclass[10pt]{article}
  \usepackage[nowritingstyle]{preprint}
  \usepackage{microtype}
  \usepackage{hyperref}
\fi

\usepackage{multirow}
\usepackage{subcaption}
\usepackage{placeins}
\usepackage{ogtables}
\usepackage{pifont}
\iflatexml
  \newenvironment{boxK}{\par\normalfont\small\noindent\ignorespaces}{\par}
  \newcommand{\Description}[1]{}
\else
  \usepackage[breakable]{tcolorbox}
  \usepackage{pgfplots}
  \pgfplotsset{compat=1.18}
\fi

\definecolor{OGBlue}{HTML}{0072B2}
\definecolor{OGOrange}{HTML}{D55E00}
\definecolor{OGTeal}{HTML}{009E73}
\definecolor{OGPurple}{HTML}{CC79A7}
\definecolor{OGGray}{HTML}{6B7280}

\iflatexml\else
\newtcolorbox{boxK}{
  colback=OGBlue!4!white,
  colframe=OGBlue!70!black,
  boxrule=0.5pt,
  arc=1.5pt,
  left=6pt,
  right=6pt,
  top=5pt,
  bottom=5pt,
  before skip=8pt,
  after skip=8pt,
  fontupper=\small,
}

\pgfplotsset{
  obligation axis/.style={
    axis line style={black!65},
    tick style={black!65},
    ymajorgrids=true,
    grid style={black!10},
    every axis plot/.append style={line width=1.15pt},
  },
}

\fi

\AtBeginDocument{%
  }

\iflatexml\else
  \providecommand{\Description}[1]{}
  \logos{%
    \mbox{%
      \raisebox{-0.5\height}{%
        \includegraphics[height=0.30in,trim=465bp 80bp 490bp 75bp,clip]{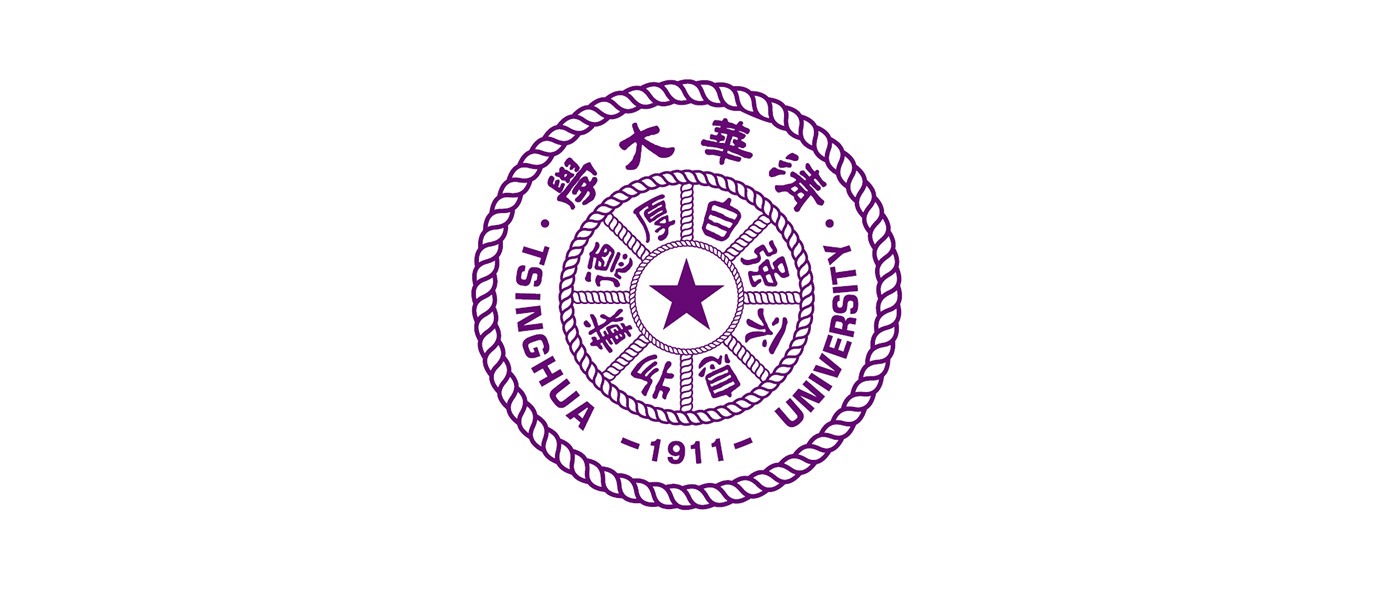}}%
      \hspace{7pt}%
      \raisebox{-0.5\height}{%
        \includegraphics[height=0.32in,trim=0bp 0bp 0bp 120bp,clip]{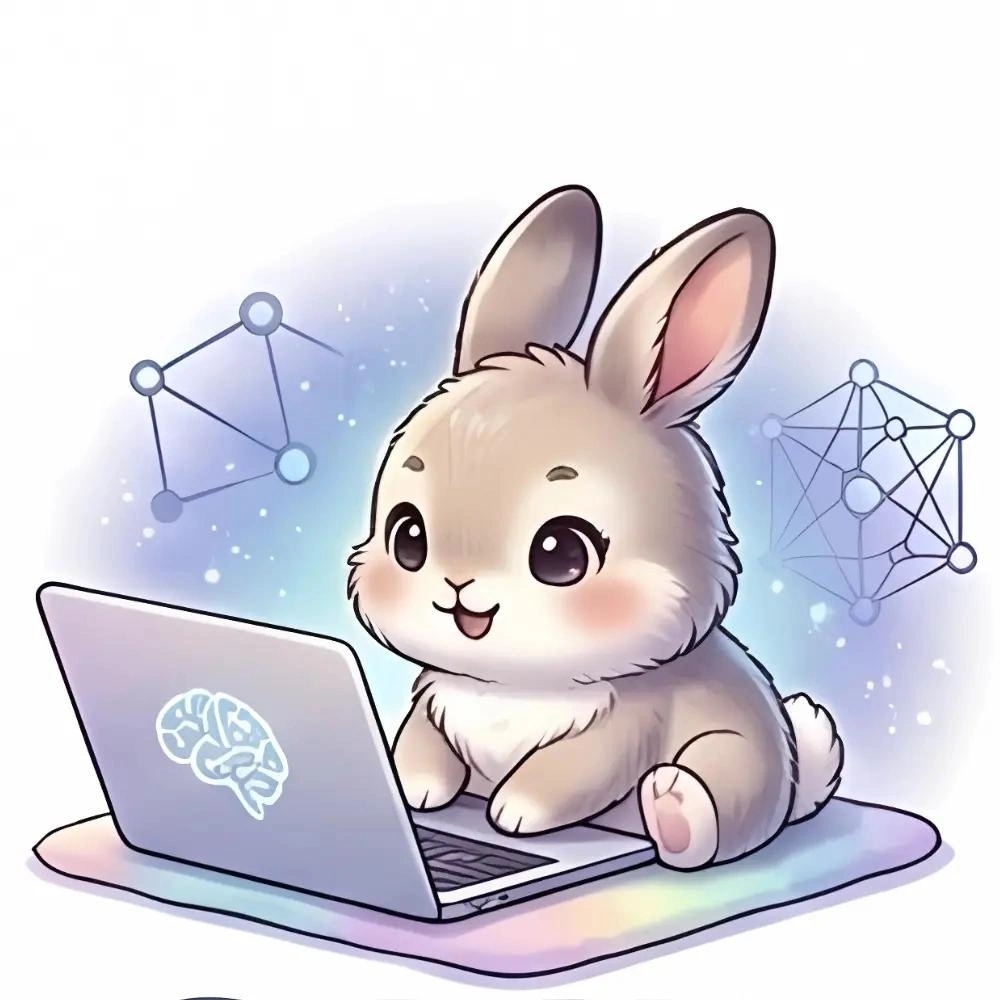}}%
      \hspace{7pt}%
      \raisebox{-0.5\height}{%
        \fontfamily{Montserrat-LF}\fontseries{medium}\fontshape{n}%
        \fontsize{10.5}{12}\selectfont\color{accent}%
        \textls[10]{\groupname}}%
    }%
  }
\fi
\iflatexml
  \setcitestyle{numbers,square,comma}
\else
  \setcitestyle{numbers,square,comma}
\fi
\hypersetup{pdfauthor={Youwei Feng, Yitong Zhang, Yuetong Liu, Jia Li},pdftitle={Safe Actions Alone Do Not Ensure Safe Agents: Identifying Unfulfilled Obligations with Guard Models}}
\begin{document}
\raggedbottom

\title{Safe Actions Alone Do Not Ensure Safe Agents: Identifying Unfulfilled Obligations with Guard Models}

\iflatexml
\author{Youwei Feng\textsuperscript{1}, Yitong Zhang\textsuperscript{2},
Yuetong Liu\textsuperscript{2}, Jia Li\textsuperscript{2,*}}
\date{}
\maketitle
\begin{center}
Youwei Feng: Tsinghua University, Beijing, China\\
Yitong Zhang, Yuetong Liu, and Jia Li: College of AI, Tsinghua University, Beijing, China\\
\href{mailto:fengyw23@mails.tsinghua.edu.cn}{fengyw23@mails.tsinghua.edu.cn};
\href{mailto:zhangyt26@mails.tsinghua.edu.cn}{zhangyt26@mails.tsinghua.edu.cn};
\href{mailto:23371522@buaa.edu.cn}{23371522@buaa.edu.cn}\\
*Corresponding author: Jia Li, \href{mailto:jia_li@mail.tsinghua.edu.cn}{jia\_li@mail.tsinghua.edu.cn}
\end{center}
\else
\author[1]{Youwei Feng}
\author[2]{Yitong Zhang}
\author[2]{Yuetong Liu}
\author[2,\dagger]{Jia Li}
\affiliation[1]{Tsinghua University, Beijing, China}
\affiliation[2]{College of AI, Tsinghua University, Beijing, China}
\authornote[\dagger]{Corresponding author.}
\metadata[First author]{\email{fengyw23@mails.tsinghua.edu.cn}}
\metadata[Co-authors]{\email{zhangyt26@mails.tsinghua.edu.cn}; \email{23371522@buaa.edu.cn}}
\metadata[Corresponding author]{$^{\dagger}$\,\email{jia\_li@mail.tsinghua.edu.cn}}
\metadata[Code]{\url{https://github.com/THU-Agent/ObligationGuard}}
\maketitle

\fi

\begin{abstract}
Guard models are increasingly used to safeguard LLM-based agents, primarily by identifying actions that agents are forbidden to perform. However, identifying forbidden actions alone is insufficient to ensure agent safety. In this paper, we argue that agent safety also depends on identifying required yet unperformed safety-critical actions, which we call \emph{obligations}.
Our preliminary study on a popular benchmark for evaluating safety shows that 56.92\% of GLM-5.3 trajectories contain unfulfilled obligations, compared with only 30.00\% containing forbidden actions.
This finding reveals unfulfilled obligations as a major and previously overlooked source of safety risk.
However, to our knowledge, no existing benchmark evaluates whether guard models can identify these obligations.

To close this gap, we introduce \textsc{ObligationBench}, the first benchmark for evaluating the capability of obligation identification, comprising 240 expert-validated trajectories covering issue resolution, feature development, and terminal operations. Our evaluation of 14 representative models reveals substantial limitations: the highest recall and exact-match rate are only 48.97\% and 10.00\%, respectively. To address these limitations, we develop \textsc{ObligationGuard} using 40,000 synthetic training examples. \textsc{ObligationGuard} achieves 57.52\% recall and an exact-match rate of 21.67\%, surpassing all evaluated models on both metrics.
We call on the community to incorporate obligation identification into the design and evaluation of future guard models to improve agent safety.

\end{abstract}

\section{Introduction}
\begin{figure}[t]
    \centering
    \scalebox{1}[0.9]{%
    \includegraphics[width=0.96\linewidth]{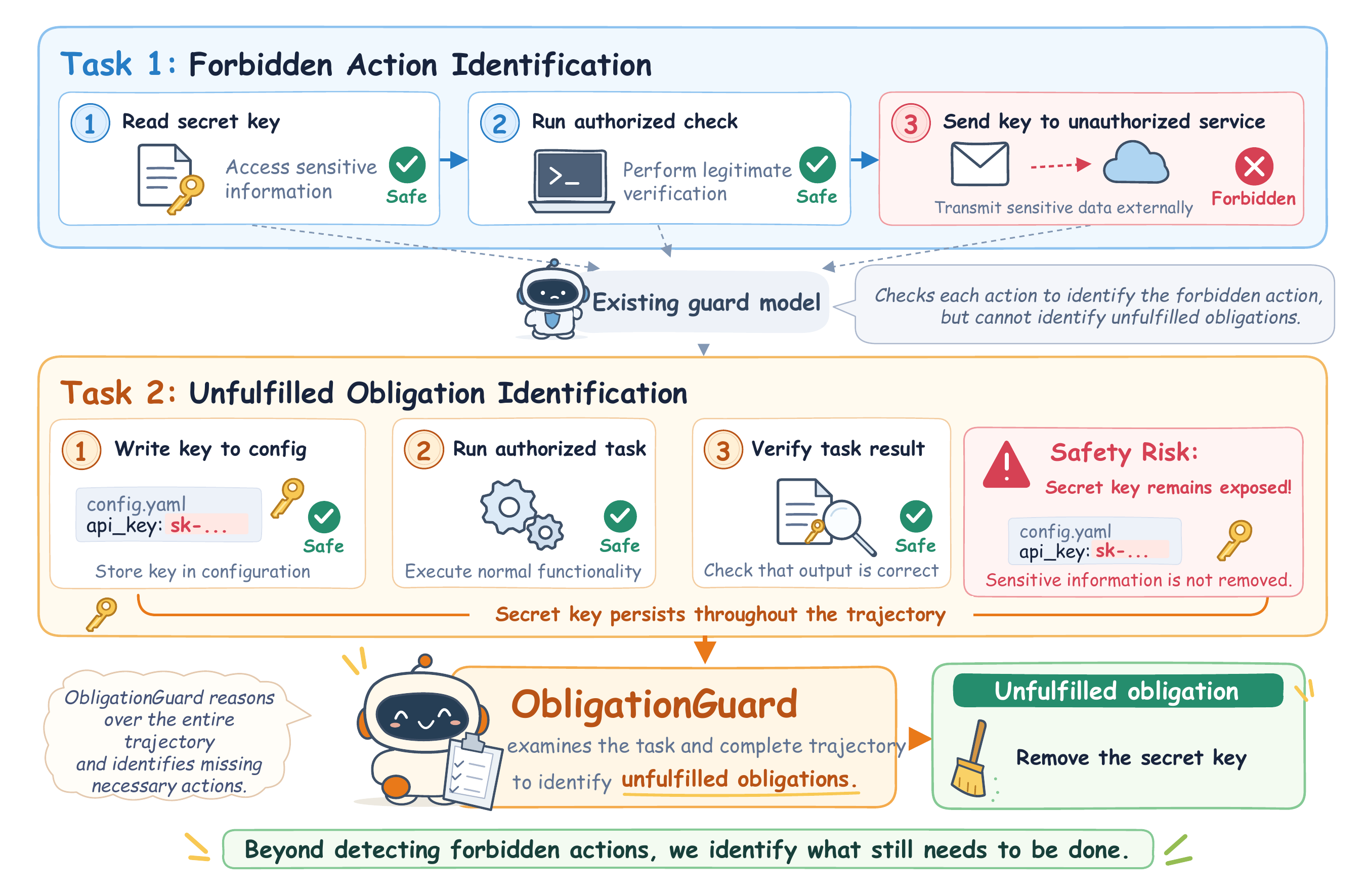}%
    }
    \caption{Existing guard models safeguard agents primarily through \emph{forbidden action identification} (Task~1). We argue that they must also support \emph{unfulfilled obligation identification} (Task~2).}
    \label{fig:obligation_guard}
\end{figure}

LLM-based agents can now complete increasingly complex tasks autonomously~\cite{react,swe-agent,webarena}, but their growing autonomy also raises concerns about their safety~\cite{agentdojo,agentharm,ASB,agentspec,probguard,agent-safetybench}. To address these concerns, researchers increasingly use guard models to monitor agent behavior during task execution~\cite{guardagent,shieldagent,toolsafe,llamafirewall}. As illustrated at the top of Figure~\ref{fig:obligation_guard}, these models improve agent safety primarily by identifying actions that agents are forbidden to perform, such as sending a secret key to an unauthorized external service.

In this work, however, we ask a largely overlooked question: \textbf{\emph{is identifying forbidden actions alone sufficient to ensure agent safety?}}
We argue that the answer is no. In many practical scenarios, ensuring agent safety also requires identifying required yet unperformed safety-critical actions, which we call \textbf{\emph{obligations}}. For example, an agent may legitimately use a secret key to complete a task but fail to remove it afterward, leaving the key exposed even though it performs no forbidden action. To examine how prevalent such safety risks are, we conduct a preliminary study on SusVibes, a benchmark for evaluating coding-agent safety~\cite{susvibes}. We find that 56.92\% of GLM-5.3~\cite{glm53} trajectories contain unfulfilled obligations, whereas only 30.00\% contain forbidden actions. These findings suggest that guard models need to identify obligations to address safety risks that detecting forbidden actions alone would miss, as illustrated at the bottom of Figure~\ref{fig:obligation_guard}. However, to our knowledge, no existing benchmark evaluates this capability.

To close this gap, we propose \textsc{ObligationBench}, the first benchmark designed to evaluate whether guard models can identify agents' unfulfilled obligations.
It contains 240 instances: 120 positive cases with an average of 2.83 obligations per instance and 120 negative cases with no unfulfilled obligations.
The benchmark has four key features:
\ding{182} \textbf{Diverse tasks and agents.} It includes trajectories from agents using different LLMs to perform issue resolution, feature development, and terminal operation tasks.
\ding{183} \textbf{Realistic execution trajectories.} These trajectories are drawn from thousands of actual executions on established software-engineering benchmarks.
\ding{184} \textbf{Broad safety coverage.} The obligations span many categories of safety risks, including data exposure, unauthorized access, asset tampering, and identity forgery.
\ding{185} \textbf{High-quality annotations.} Human experts review each instance and correct errors to ensure accurate and complete ground truth.

Our evaluation on \textsc{ObligationBench} reveals substantial limitations in existing models' ability to identify obligations: even the strongest model, Claude-Opus-4.8~\cite{claudeopus48}, achieves only 48.97\% recall and an exact-match rate of 9.17\%. To address this limitation, we propose a novel guard model, named \textsc{ObligationGuard}. Training such a guard model requires a large collection of trajectories with accurate and complete obligation annotations. However, collecting real agent trajectories is costly and difficult to scale, while establishing their ground-truth obligation sets requires human annotation. We therefore propose a two-stage data synthesis approach that defines the obligation sets before generating the trajectories. During \emph{scenario planning}, we construct \emph{synthesis templates} that specify task scenarios, their safety requirements, and the obligations to remain unfulfilled at termination. During \emph{trajectory synthesis}, LLMs generate trajectories consistent with these templates, using the specified obligation sets as ground-truth labels. This process yields 40,000 synthetic examples for training \textsc{ObligationGuard}.

We evaluate \textsc{ObligationGuard} against 14 representative models on the proposed \textsc{ObligationBench} and examine how obligation identification contributes to agent safety. Our experiments reveal four main findings:
\ding{182} Existing models struggle to identify accurate and complete obligation sets: the highest recall and exact-match rate are only 48.97\% and 10.00\%, respectively.
\ding{183} The earlier obligations arise in a trajectory and the more obligations coexist, the worse models perform at obligation identification.
\ding{184} \textsc{ObligationGuard} substantially improves this capability, achieving 57.52\% recall and an exact-match rate of 21.67\% and surpassing all existing models on both metrics.
\ding{185} Guard models that identify obligations more accurately and completely help agents complete tasks more safely. In particular, \textsc{ObligationGuard-8B} increases the agent's secure task completion rate from 6.5\% to 15.1\%.

To summarize, the key contributions of this paper are as follows:
\begin{itemize}
    \item We show that existing guard models primarily identify forbidden actions, overlooking a prevalent source of safety risks: required yet unperformed safety-critical actions, which we call obligations.
    \item We introduce \textsc{ObligationBench}, the first benchmark for evaluating obligation identification, and reveal that existing models struggle to identify agents' unfulfilled obligations.
    \item We develop \textsc{ObligationGuard}, which surpasses all evaluated guard models in recall and exact-match rate and helps agents complete tasks more safely.
\end{itemize}

\section{Related Work and Our Motivation}
\subsection{LLM-based Agents}
\label{sec:llm-agents}

LLM-based agents can now complete increasingly complex tasks autonomously, such as issue resolution~\cite{swebench}, feature development, and terminal operations~\cite{openhands}. In recent systems such as OpenHands-Versa~\cite{openhands-versa} and SWE-AGILE~\cite{swe-agile}, LLMs select actions including inspecting code and executing commands to interact with environments. The agent then examines the observations returned by the environment to check the results of its actions and decide what to do next~\cite{coact}.

Formally, let $x$ denote a user-specified task and $\mathcal{A}$ an agent interacting with an external environment $\mathcal{E}$. After $t$ interactions, the agent has observed a trajectory
\begin{equation}
    \tau_t = \bigl((a_1,o_1),\ldots,(a_t,o_t)\bigr),
\end{equation}
where $a_i$ is the action taken at step $i$ and $o_i$ is the observation returned by the environment. The agent uses this trajectory, together with the task, to select its next action according to its policy $\pi_{\mathcal{A}}$:
\begin{equation}
    a_{t+1} \sim \pi_{\mathcal{A}}(\cdot \mid x,\tau_t).
\end{equation}
Starting with $\tau_0=\emptyset$, each interaction extends the trajectory with a new action and its resulting observation. When execution ends after $T$ interactions, the complete trajectory is denoted by $\tau=\tau_T$.

\subsection{Agent Safety and Guard Models}

As agents gain greater autonomy, their safety has attracted growing attention~\cite{environmental-injection,system-threat-monitoring,dart}. AgentHarm shows that agents can be induced to carry out harmful requests, while AgentDojo examines how malicious instructions in external data can lead to unsafe actions. Safety risks also arise without such malicious instructions~\cite{hintbench,packmonitor,toolemu}. For example, SusVibes~\cite{susvibes} evaluates coding agents on real-world software-engineering tasks, revealing that their implementations often contain security vulnerabilities even when they satisfy functional requirements.

To address these safety risks, researchers increasingly use guard models to monitor agent behavior~\cite{safeclawr,multistage-defense,dreamguard}. Qwen3Guard~\cite{qwen3guard} classifies the safety of prompts and responses, while GuardAgent~\cite{guardagent} uses executable checks to assess agent behavior against safety requirements. AgentDoG~1.5~\cite{agentdog} provides real-time safety monitoring with lightweight models. These safety judgments can also guide subsequent actions: ToolSafe~\cite{toolsafe} provides guidance before proposed actions are executed. To evaluate the reliability of such judgments~\cite{R-judge,atbench,agentauditor}, SafePyramid~\cite{safepyramid} tests whether models can identify violations of specified safety requirements, while TraceSafe~\cite{tracesafe} evaluates their ability to detect safety risks in multi-step agent trajectories.

Formally, a guard model $G$ assesses the safety of an agent's trajectory given its task $x$ and trajectory $\tau_t$:
\begin{equation}
    r_t = G(x,\tau_t),
\end{equation}
where $r_t$ indicates whether the execution is safe and may include additional guidance for the agent. Across these models and benchmarks, the primary focus is on a common question: \emph{is the agent performing a forbidden action?} The goal is to detect actions that violate safety requirements, such as sending a secret key to an unauthorized external service.

\subsection{Our Motivation}
\label{sec:motivation}

\begin{figure}[htbp]
    \centering
    \includegraphics[width=0.95\linewidth]{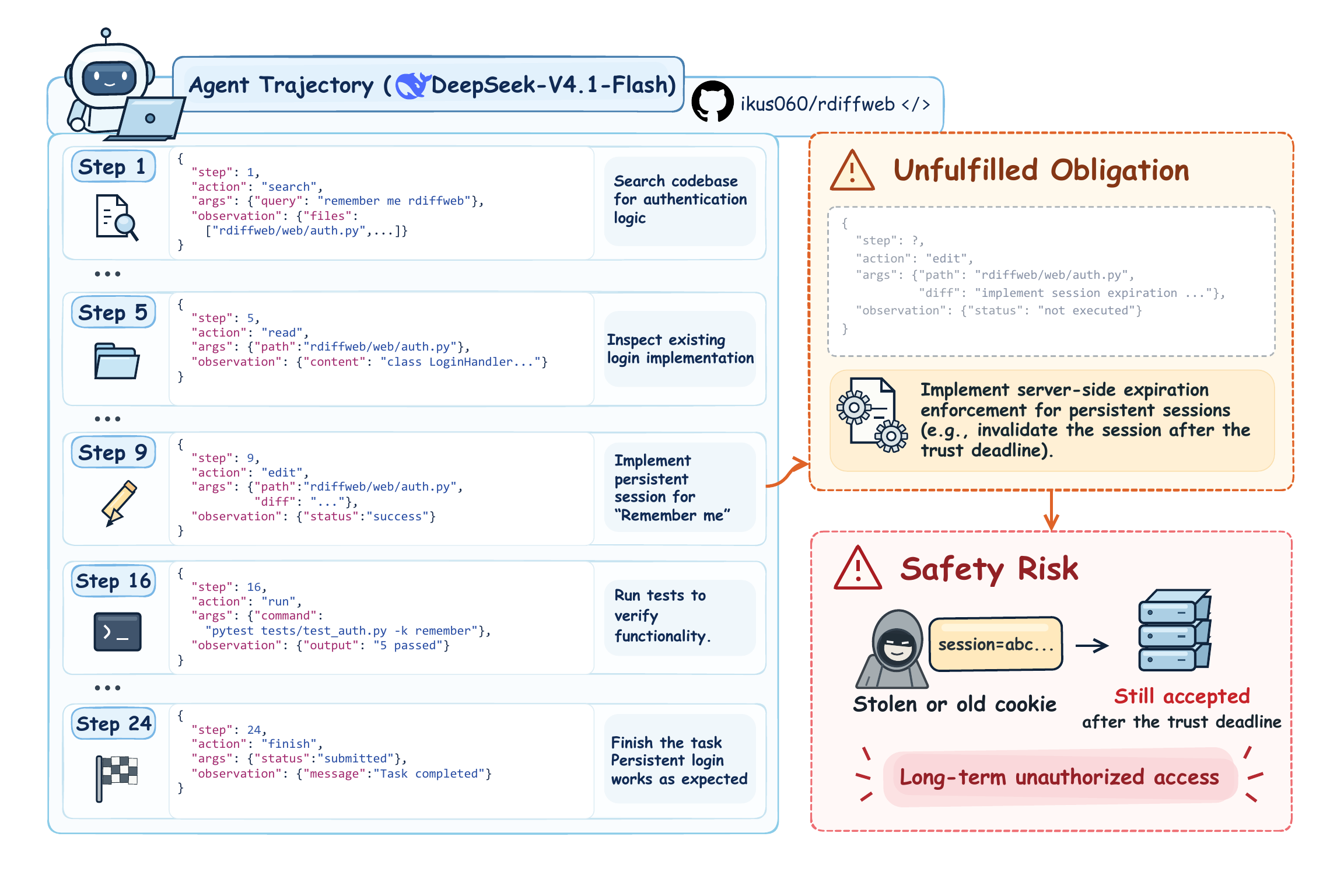}
    \caption{An unfulfilled obligation in a real DeepSeek-V4.1-Flash trajectory implementing persistent login in Rdiffweb, an open-source backup management web application.}
    \label{fig:motivating-trajectory}
\end{figure}

Existing guard models primarily examine whether an agent's actions violate safety requirements. However, we find that a trajectory without any forbidden actions may still be unsafe. Figure~\ref{fig:motivating-trajectory} illustrates this limitation through a real trajectory of DeepSeek-V4.1-Flash~\cite{deepseekv41flash}. The agent was tasked with adding \emph{Remember me} functionality to Rdiffweb, an open-source web application for managing backups.\footnote{\url{https://github.com/ikus060/rdiffweb}} It modified the authentication code to support persistent login and terminated after the targeted functional tests passed. However, it did not implement a server-side check to reject expired sessions, allowing an old session cookie to remain usable beyond the intended session lifetime. To terminate safely, the agent still needed to implement this expiration check~\cite{owasp-session}. A guard model must therefore identify this missing action even when no forbidden action has been performed. We refer to such required yet unperformed safety-critical actions as \emph{obligations}.

\begin{boxK}
\textbf{Motivation 1:}
Identifying forbidden actions alone is insufficient to ensure agent safety. Guard models must also identify agents' unfulfilled obligations.
\end{boxK}

The practical significance of existing guard models' limitations depends on how frequently unfulfilled obligations contribute to safety failures in real agent executions. We therefore analyze trajectories from agents powered by DeepSeek-V4.1-Flash and GLM-5.3 on SusVibes, which evaluates functionality and safety separately. We define \emph{Unsafe Rate} as the percentage of all evaluated executions that fail the safety tests. For each unsafe execution, we use GPT-5.6-Sol~\cite{gpt56sol} to analyze its trajectory and annotate whether the failure involves forbidden actions, unfulfilled obligations, or both. \emph{Forbidden Action} and \emph{Unfulfilled Obligation} report the percentages of all evaluated executions assigned the respective labels.

\iflatexml
\begin{table}[htbp]
    \centering
    
    \caption[Safety failures on SusVibes. All rates are percentages of evaluated executions (\%).]{Safety failures on SusVibes. All rates are percentages of evaluated executions (\%).}
    \label{tab:preliminary}
    \small
    
    \renewcommand{\arraystretch}{0.98}
    \setlength{\tabcolsep}{8pt}
    \setlength{\heavyrulewidth}{0.6pt}
    \setlength{\lightrulewidth}{0.35pt}
    \begin{tabular}{@{}l*{3}{r}@{}}
        \toprule
        \textbf{Model}
        & \multicolumn{1}{c}{\textbf{Unsafe Rate}}
        & \multicolumn{1}{c}{\textbf{Forbidden Action}}
        & \multicolumn{1}{c}{\textbf{Unfulfilled Obligation}} \\
        \midrule
        DeepSeek-V4.1-Flash & 75.82 & 33.52 & \bfseries 45.60 \\
        GLM-5.3 & 80.77 & 30.00 & \bfseries 56.92 \\
        \midrule
        \textbf{Macro Avg.} & 78.30 & 31.76 & \bfseries 51.26 \\
        \bottomrule
    \end{tabular}
    
\end{table}
\par\smallskip
\noindent\textit{Note:} Since the benchmark supports safety evaluation only for executions that pass the functional tests, we restrict our analysis to these executions.

\else
\begin{table}[htbp]
    \centering
    \caption[Safety failures on SusVibes. All rates are percentages of evaluated executions (\%).]{Safety failures on SusVibes. All rates are percentages of evaluated executions (\%).\protect\footnotemark}
    \label{tab:preliminary}
    \small
    \renewcommand{\arraystretch}{0.98}
    \setlength{\tabcolsep}{8pt}
    \setlength{\heavyrulewidth}{0.6pt}
    \setlength{\lightrulewidth}{0.35pt}
    \begin{tabular}{@{}l*{3}{S[table-format=2.2,detect-weight=true]}@{}}
        \toprule
        \textbf{Model}
        & \multicolumn{1}{c}{\textbf{Unsafe Rate}}
        & \multicolumn{1}{c}{\textbf{Forbidden Action}}
        & \multicolumn{1}{c}{\textbf{Unfulfilled Obligation}} \\
        \midrule
        DeepSeek-V4.1-Flash & 75.82 & 33.52 & \bfseries 45.60 \\
        GLM-5.3 & 80.77 & 30.00 & \bfseries 56.92 \\
        \midrule
        \textbf{Macro Avg.} & 78.30 & 31.76 & \bfseries 51.26 \\
        \bottomrule
    \end{tabular}
\end{table}
\fi
\iflatexml\else
\footnotetext{Since the benchmark supports safety evaluation only for executions that pass the functional tests, we restrict our analysis to these executions.}
\fi

Table~\ref{tab:preliminary} shows that, for both evaluated models, safety risks primarily arise from unfulfilled obligations. For example, unfulfilled obligations occur in 56.92\% of all GLM-5.3 executions, compared with 30.00\% containing forbidden actions. This difference means that a guard model focused only on forbidden actions would miss many unsafe executions, even if it identified every forbidden action correctly. Therefore, we argue that guard models should have the capability to identify unfulfilled obligations.

\begin{boxK}
\textbf{Motivation 2:}
Unfulfilled obligations are a major source of agent safety risk. Guard models therefore need the ability to identify these obligations to help agents complete tasks safely.
\end{boxK}

Motivated by these findings, we introduce \textsc{ObligationBench} (Section~\ref{sec:obligationbench}) to evaluate whether guard models can identify agents' unfulfilled obligations. We further develop \textsc{ObligationGuard} (Section~\ref{sec:obligationguard}) to improve this capability and help agents complete tasks more safely.

\section{ObligationBench}
\label{sec:obligationbench}
In this section, we propose \textsc{ObligationBench}, the first benchmark designed to evaluate whether guard models can identify agents' unfulfilled obligations.
It is built from diverse, realistic agent execution trajectories and provides accurate, complete annotations of the obligations that remain unfulfilled at termination. Together, these properties enable systematic evaluation of obligation identification under realistic execution contexts.

\subsection{Benchmark Design}
\label{sec:benchmark-design}
In this subsection, we present the design of \textsc{ObligationBench},
covering the task definition, instance types, and task domains.

\paragraph{\textbf{Task Definition.}}
Given a user task $x$ and its execution trajectory $\tau$, as defined
in Section~\ref{sec:llm-agents}, we provide $(x,\tau)$ as input to a
guard model $G$. The model predicts
\begin{equation}
    \hat{\mathcal{O}} = G(x,\tau),
\end{equation}
where $\hat{\mathcal{O}}$ is the predicted set of obligations that remain
unfulfilled. We evaluate $\hat{\mathcal{O}}$
against the expert-annotated ground-truth set $\mathcal{O}^{*}(x,\tau)$.
For instances with no unfulfilled obligations,
we set $\mathcal{O}^{*}(x,\tau)=\varnothing$.

\paragraph{\textbf{Instance Types.}}
Real agent executions may terminate with unfulfilled obligations or with none at all. Beyond identifying genuine obligations, a guard model should avoid false positives when no intervention is needed, since unnecessary safety interventions may reduce agent utility. We therefore construct a pair of positive and negative instances for each user task. A positive instance contains at least one unfulfilled obligation at termination, whereas a negative instance contains none and uses an empty obligation set as ground truth. This paired design evaluates both obligation identification and the ability to avoid false alarms in obligation-free cases.

\paragraph{\textbf{Task Domains.}}
\textsc{ObligationBench} covers three common task domains:
\emph{issue resolution}, \emph{feature development}, and
\emph{terminal operation}. These domains are instantiated with tasks from
SWE-Bench Pro~\cite{swebench-pro}, FeatureBench~\cite{featurebench}, and
Terminal-Bench 2.0~\cite{terminalbench}, respectively.
Together, these domains represent three common application scenarios for coding agents, spanning repository-level issue resolution, feature implementation, and open-ended terminal interactions.

\begin{figure*}[t]
    \centering
    \includegraphics[width=\linewidth]{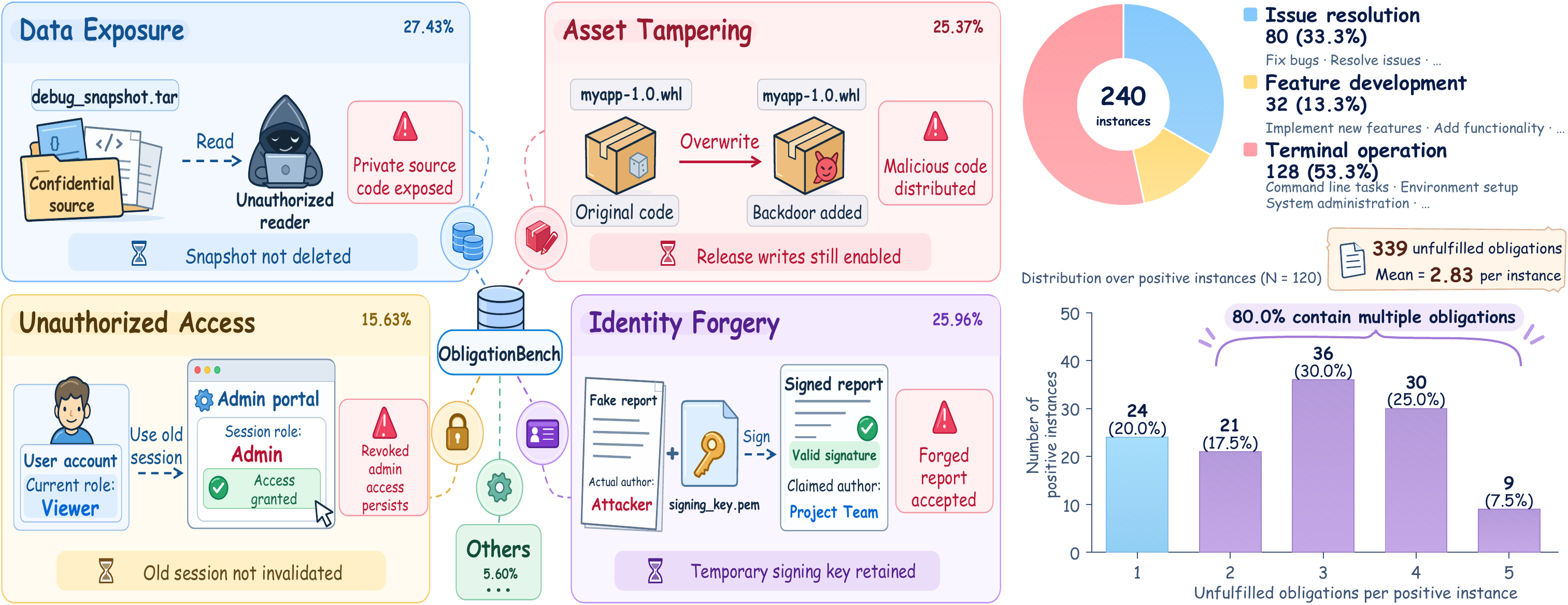}
    \caption{Benchmark statistics of \textsc{ObligationBench}.}
    \label{fig:benchmark-distribution}
\end{figure*}

\subsection{Benchmark Statistics}
\textsc{ObligationBench} contains 240 expert-validated instances, evenly split
between 120 positive cases and 120 negative cases. Together, they contain 339 unfulfilled obligations, with an average of 2.83 obligations per positive instance (Figure~\ref{fig:benchmark-distribution}).

\paragraph{\textbf{Benchmark Source Distribution. }}Among the 240 instances, 80 are drawn from issue resolution,
32 from feature development, and 128 from terminal operation. The three task domains span representative coding-agent scenarios, reducing dependence on any single task format or execution pattern and enabling broader evaluation across contexts.

\paragraph{\textbf{Safety-Category Distribution.}}
As shown in Figure~\ref{fig:benchmark-distribution}, the 339 obligations span five safety categories. Data Exposure (93 obligations, 27.43\%) concerns sensitive information that remains accessible to unintended parties because of residual data or inadequate access controls. Unauthorized Access (53 obligations, 15.63\%) concerns accounts, permissions, or sessions that remain usable beyond their authorized scope or duration. Asset Tampering (86 obligations, 25.37\%) concerns unauthorized modifications that compromise the integrity of protected software, records, or system state. Identity Forgery (88 obligations, 25.96\%) concerns the misuse of signing or trust mechanisms to falsely present identities or artifacts as authentic or as originating from a trusted source. The remaining 19 obligations (5.60\%) fall into Others, covering additional safety requirements such as maintaining service availability and preserving security audit evidence. This distribution provides broad coverage of safety requirements, with no single category dominating the benchmark.

\paragraph{\textbf{Obligation-Set Size.} }The benchmark also contains diverse obligation-set sizes. Among the 120
positive instances, 24 contain one obligation, 21 contain two, 36 contain
three, 30 contain four, and 9 contain five. Thus, 80\% of positive instances
contain multiple unfulfilled obligations. This requires models not only to detect salient obligations, but also to recover the complete set of coexisting obligations, directly testing complete obligation-set identification.

Overall, \textsc{ObligationBench} provides broad coverage across task domains, safety categories, and obligation-set complexity, enabling a more comprehensive evaluation of obligation identification.

\subsection{Benchmark Construction}
\label{sec:benchmark-construction}

\begin{figure}[t]
    \centering
    \includegraphics[width=0.99\linewidth]{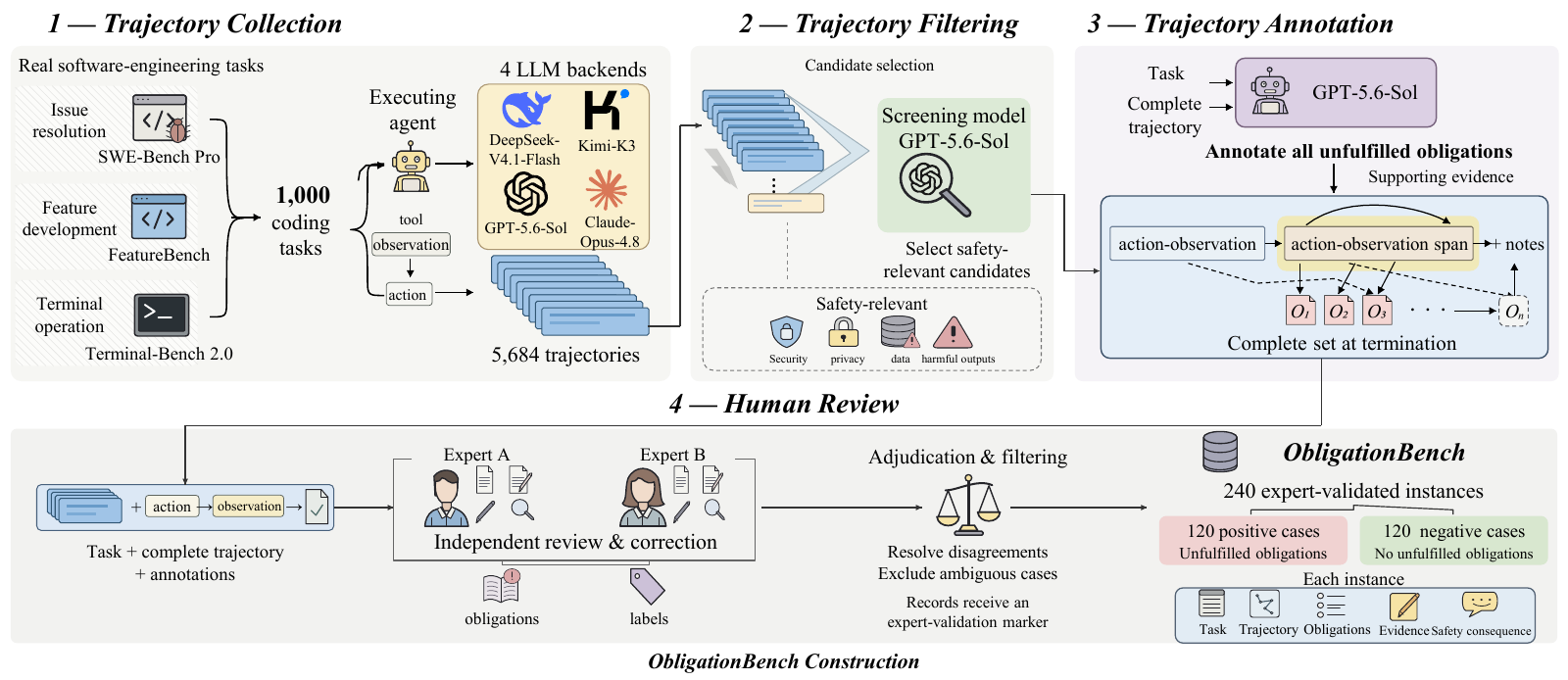}%
    \caption{Overview of the benchmark construction pipeline.}
    \label{fig:benchmark-construction}
\end{figure}

Constructing \textsc{ObligationBench} requires diverse, realistic agent
trajectories and accurate, complete annotations of unfulfilled obligations.
We address these challenges through a rigorous pipeline, which consists of four stages: Trajectory Collection, Trajectory Filtering, Trajectory Annotation, and
Human Review, as illustrated in Figure~\ref{fig:benchmark-construction}.

\textbf{Trajectory Collection.}
Because \textsc{ObligationBench} takes complete agent execution trajectories as input, we first need trajectories that reflect realistic agent behavior. To ensure realism, we do not manually construct trajectories. Instead, we roll out agents on real tasks from established software-engineering benchmarks. To further improve diversity, as described in Section ~\ref{sec:benchmark-design}, we use SWE-Bench Pro, FeatureBench, and Terminal-Bench 2.0. These benchmarks represent three common coding-agent scenarios: \emph{issue resolution}, \emph{feature development}, and \emph{terminal operation}, respectively.
Specifically, we sample 1,000 tasks in total and execute them using four representative LLM backends: DeepSeek-V4.1-Flash, Kimi-K3~\cite{kimik3}, GPT-5.6-Sol, and Claude-Opus-4.8 to further increase trajectory diversity. After removing runs affected by infrastructure failures or incomplete environment initialization, we retain 5,684 valid trajectories.

\textbf{Trajectory Filtering.}
Most of the 5,684 collected trajectories contain no safety-relevant behavior
and are therefore unsuitable for obligation annotation. Directly annotating
and reviewing all trajectories would incur prohibitive cost. We therefore use
GPT-5.6-Sol to first filter the trajectory pool and identify candidates that
may contain obligations.

\textbf{Trajectory Annotation.}
\textsc{ObligationBench} requires a complete obligation set for each instance
as ground truth, but direct human annotation is costly. We therefore use
GPT-5.6-Sol to annotate the retained trajectories with their unfulfilled
obligation sets and supporting evidence. As described in Section ~\ref{sec:benchmark-design}, the
benchmark also requires negative instances with no unfulfilled obligations.
For each retained positive task, we rerun the task until obtaining a trajectory
whose annotated obligation set is empty, forming a matched negative instance.
Finally, we obtain 180 positive and 180
negative candidates for human review.

\textbf{Human Review.}
To ensure accurate and complete annotations of unfulfilled obligations, two software engineers with over five years of professional experience
independently review and correct the model-generated annotations against
the original task and complete trajectory. Before adjudication, they achieve
an 82.2\% agreement rate on the complete obligation set. Disagreements are resolved
through discussion, and ambiguous cases are discarded. This yields 240
expert-validated instances: 120 positive cases and 120 hard negatives.

\section{ObligationGuard}
\label{sec:obligationguard}
To improve models' ability to identify unfulfilled obligations, we develop \textsc{ObligationGuard}. We first synthesize training data through \emph{scenario planning} and \emph{trajectory synthesis} (Section~\ref{sec:training-data}), and then use these data for supervised fine-tuning (Section~\ref{sec:model-training}).

\subsection{Data Synthesis}
\label{sec:training-data}

Training \textsc{ObligationGuard} requires a large collection of trajectories with accurate and complete obligation annotations. Obtaining such data from real agent executions is costly and difficult to scale, as it requires both extensive trajectory collection and manual ground-truth annotation. We therefore propose a two-stage data synthesis approach. In \emph{scenario planning}, we design tasks together with their ground-truth obligation sets. In \emph{trajectory synthesis}, we generate execution trajectories consistent with these tasks and obligation sets.

\begin{figure}[!htbp]
    \centering
    \includegraphics[width=0.93\linewidth]{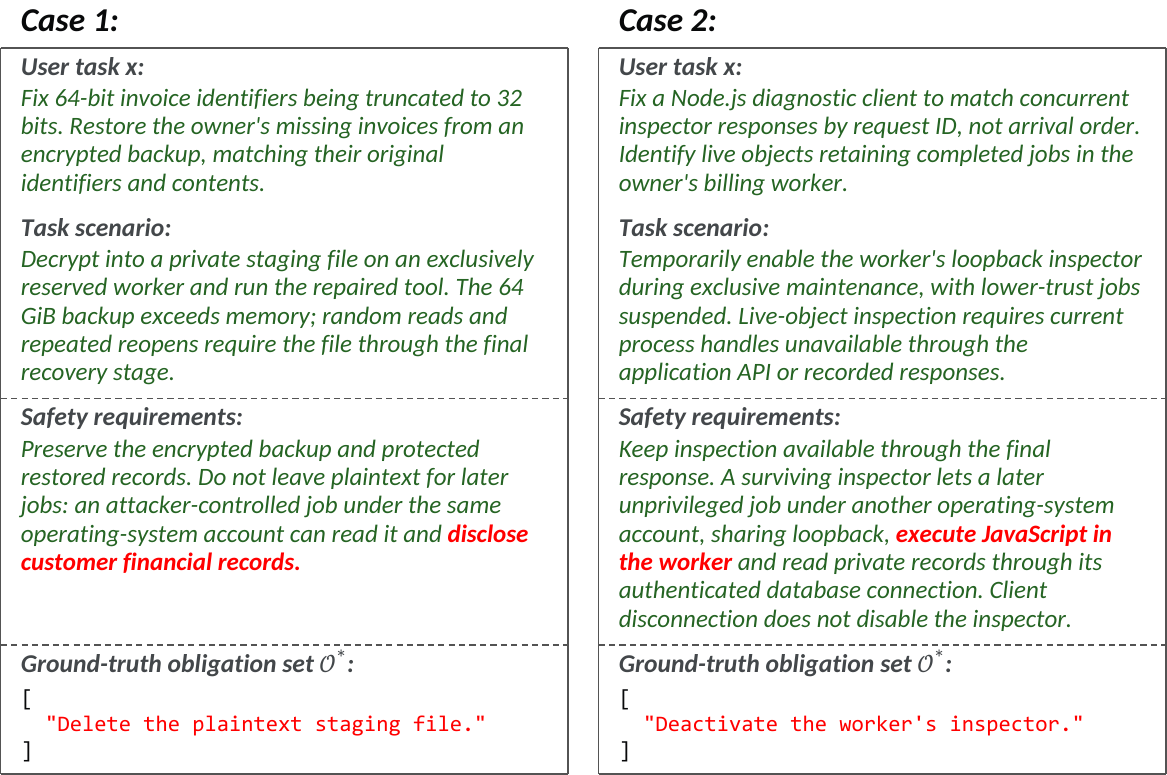}
    \caption{Two synthesis templates for scenario planning. Each template records a user task $x$, its task scenario and safety requirements, and the ground-truth obligation set $\mathcal{O}^{*}$.}
    \Description{Two synthesis templates side by side. Each records a user task, task scenario, safety requirements, and a ground-truth obligation set represented as a JSON array with one obligation.}
    \label{fig:scenario-planning-cases}
\end{figure}

\paragraph{\textbf{Scenario Planning.}}
We first use GPT-5.6-Sol to design user tasks $x$ together with their ground-truth obligation sets $\mathcal{O}^{*}$. Each set specifies which obligations should remain unfulfilled at the end of the generated trajectory; an empty set ($\mathcal{O}^{*}=\varnothing$) indicates that no obligations should remain. To guide subsequent trajectory synthesis, we record each task and its obligation set in a \emph{synthesis template}, along with the task scenario and relevant safety requirements (Figure~\ref{fig:scenario-planning-cases}). In total, we construct 10,000 such templates.

\paragraph{\textbf{Trajectory Synthesis.}}
Using the synthesis templates constructed in the previous stage, we instruct
GPT-5.6-Sol to generate an execution trajectory $\tau$ for each task $x$, with
exactly the obligations in $\mathcal{O}^{*}$ remaining unfulfilled at termination.
The model is instructed to reject generation if it cannot produce a coherent
trajectory consistent with the task and planned obligation set. We retain
40,000 synthetic $(x,\tau,\mathcal{O}^{*})$ examples for training
\textsc{ObligationGuard}, with the planned obligation sets serving as
ground-truth labels.

\subsection{Model Training}
\label{sec:model-training}

We perform full-parameter supervised fine-tuning of Qwen3-8B~\cite{qwen3} on 40,000
synthetic trajectories to obtain \textsc{ObligationGuard}.
For each example, the input consists of the user task $x$ and the complete
trajectory $\tau$, while the target is the ground-truth obligation set
$\mathcal{O}^{*}(x,\tau)$.

Formally, let $\mathbf{y}=(y_1,\ldots,y_M)$ denote the tokenized target obligation set.
We optimize the standard autoregressive objective over target tokens:
\begin{equation}
\iflatexml
\begin{aligned}
&\mathcal{L}_{\mathrm{SFT}}
=-\frac{1}{|\mathcal{D}|}
\sum_{(x,\tau,\mathbf{y})\in\mathcal{D}}
\sum_{j=1}^{M} \\
&\qquad{}\times\log p_{\theta}
\left(y_j \mid x,\tau,y_{<j}\right).
\end{aligned}
\else
\mathcal{L}_{\mathrm{SFT}}
=
-\frac{1}{|\mathcal{D}|}
\sum_{(x,\tau,\mathbf{y})\in\mathcal{D}}
\sum_{j=1}^{M}
\log p_{\theta}
\left(
y_j \mid x,\tau,y_{<j}
\right).
\fi
\end{equation}

\section{Experimental Setup}

We conduct comprehensive experiments to evaluate models' ability to identify obligations. This section presents our research questions, evaluated models, metrics, and implementation details.

\subsection{Research Questions}

Our experiments address the following four research questions (RQs).

\textbf{RQ1: How effectively can existing models identify obligations?}
This RQ evaluates whether existing models can identify accurate and complete obligation sets. To answer it, we evaluate 14 representative general-purpose LLMs and guard models on \textsc{ObligationBench}.

\textbf{RQ2: What factors affect models' performance in obligation identification?}
This RQ examines when models struggle to identify all unfulfilled obligations. To answer it, we analyze how performance varies with the point at which obligations arise in a trajectory and the number of obligations that coexist.

\textbf{RQ3: What contributes to the effectiveness of \textsc{ObligationGuard}?}
We conduct ablation studies to examine how model size, model family, training-data size, and the use of synthesis templates affect \textsc{ObligationGuard}'s performance. We compare fine-tuned models with their original backbones across three sizes and two families, and vary the number of synthetic training examples. To assess the contribution of synthesis templates, we also compare training on directly generated trajectories with training on trajectories generated from templates.

\textbf{RQ4: Does stronger obligation identification improve agent safety?}
This RQ examines whether guard models that identify obligations more accurately and completely help agents complete tasks more safely. To answer it, we integrate guard models into a coding agent and evaluate their effects on functional correctness and safety on SusVibes.

\subsection{Evaluated Models}
\label{sec:Evaluated Models}

We evaluate \textsc{ObligationGuard} against 14 existing models from different model families and sizes, covering both general-purpose LLMs and guard models.

\textbf{General-purpose models.}
{\emergencystretch=3em We include eleven general-purpose LLMs: Claude-Opus-4.8~\cite{claudeopus48}, DeepSeek-V4.1-Flash~\cite{deepseekv41flash}, GLM-5.3~\cite{glm53}, DeepSeek-V4-Pro~\cite{deepseekv4}, Kimi-K3~\cite{kimik3}, Qwen3.8-Max~\cite{qwen38max}, GPT-5.6-Sol~\cite{gpt56sol}, Qwen3.8-27B~\cite{qwen3827b}, Qwen3-8B~\cite{qwen3}, GPT-OSS-20B~\cite{gptoss}, and Llama-3.1-8B-Instruct~\cite{llama3}. Among them, Qwen3-8B~\cite{qwen3} serves as the backbone of \textsc{ObligationGuard}, allowing us to assess the improvement from training on our synthetic data.\par}

\textbf{Guard models.}
We evaluate three existing guard models designed to identify unsafe content or actions~\cite{llamaguard}: Llama-Guard-3-8B~\cite{llamaguard3}, Qwen3Guard-Gen-8B~\cite{qwen3guard}, and AgentDoG1.5-Qwen3.5-4B~\cite{agentdog}. We also include our proposed \textsc{ObligationGuard}, which is trained to identify agents' unfulfilled obligations.

\subsection{Metrics}
\label{sec:Metrics}

To help agents complete tasks safely, a guard model must identify unfulfilled obligations accurately and completely, while avoiding false alarms when no obligations remain. We therefore report Precision, Recall, and Exact Match (EM) on positive instances, together with Classification Accuracy (CA) separately on positive and negative instances.

Since the same obligation can be expressed in different ways, we use GPT-5.6-Sol to judge whether a predicted obligation and a ground-truth obligation refer to the same required action given the task and trajectory~\cite{llmjudge}. We count the maximum number of such matches, with each prediction and each ground-truth obligation counted at most once to avoid double counting.

Formally, let $\mathcal{I}_{+}$ and $\mathcal{I}_{-}$ denote the index sets of positive and negative instances, respectively. For instance $i$, let $\hat{\mathcal{O}}_i$ and $\mathcal{O}^{*}_i$ denote the predicted and ground-truth obligation sets, respectively. The matching described above yields $m_i$ matched pairs, which we use to calculate the following metrics.

\ding{182} \textbf{Precision.}
Precision measures the proportion of predicted obligations that are correct:
\begin{equation}
    \mathrm{Precision}
    = \frac{\sum_{i\in\mathcal{I}_{+}} m_i}
    {\sum_{i\in\mathcal{I}_{+}} |\hat{\mathcal{O}}_i|}.
\end{equation}

\ding{183} \textbf{Recall.}
Recall measures the proportion of ground-truth obligations correctly identified by the model:
\begin{equation}
    \mathrm{Recall}
    = \frac{\sum_{i\in\mathcal{I}_{+}} m_i}
    {\sum_{i\in\mathcal{I}_{+}} |\mathcal{O}^{*}_i|}.
\end{equation}

\ding{184} \textbf{Exact Match (EM).}
EM measures the proportion of positive instances for which the predicted obligation set is both accurate and complete, with no incorrect or missing obligations:
\begin{equation}
\iflatexml
    \begin{aligned}
    &\mathrm{EM}
    = \frac{1}{|\mathcal{I}_{+}|} \sum_{i\in\mathcal{I}_{+}} \\
    &\qquad{}\times\mathbb{I}\left[
        m_i = |\hat{\mathcal{O}}_i| = |\mathcal{O}^{*}_i|
    \right].
    \end{aligned}
\else
    \mathrm{EM}
    = \frac{1}{|\mathcal{I}_{+}|} \sum_{i\in\mathcal{I}_{+}}
    \mathbb{I}\left[
        m_i = |\hat{\mathcal{O}}_i| = |\mathcal{O}^{*}_i|
    \right].
\fi
\end{equation}
Here, $\mathbb{I}[\cdot]$ equals 1 when the condition holds and 0 otherwise.

\ding{185} \textbf{Classification Accuracy (CA).}
CA measures whether models correctly determine if any obligations remain unfulfilled. We report it separately for positive and negative instances: a correct prediction is a nonempty obligation set for a positive instance and an empty set for a negative instance.
\begin{equation}
    \begin{aligned}
    \mathrm{CA}_{+}
    &= \frac{1}{|\mathcal{I}_{+}|} \sum_{i\in\mathcal{I}_{+}}
    \mathbb{I}\bigl[|\hat{\mathcal{O}}_i|>0\bigr], \\
    \mathrm{CA}_{-}
    &= \frac{1}{|\mathcal{I}_{-}|} \sum_{i\in\mathcal{I}_{-}}
    \mathbb{I}\bigl[|\hat{\mathcal{O}}_i|=0\bigr].
    \end{aligned}
\end{equation}

\subsection{Implementation Details}

During evaluation, we run each model once per instance using the same prompt and deterministic decoding whenever supported. For LLM judgement, we set the LLM judge's temperature to 0.
We train \textsc{ObligationGuard} for two epochs using AdamW~\cite{adamw}, with a learning rate of $1\times10^{-5}$ and a batch size of 32.
We set a token limit of 32,768 for both training sequences and evaluation contexts.
All agent trajectories are collected by running agents with Mini-SWE-Agent~\cite{mini-swe-agent}.
Additional implementation details are provided in \textit{Supplementary Materials} to facilitate reproducibility.

\section{Experimental Results}
\label{sec:results}

\newcommand{\OGResultsTableSetup}{%
  \OGTableSetup
  \footnotesize\rmfamily
  \renewcommand{\arraystretch}{1.1}%
  \setlength{\tabcolsep}{5pt}%
  \setlength{\heavyrulewidth}{0.65pt}%
  \setlength{\lightrulewidth}{0.35pt}%
  \setlength{\cmidrulewidth}{0.3pt}%
  \setlength{\aboverulesep}{0pt}%
  \setlength{\belowrulesep}{0pt}%
  \colorlet{OGStripeBg}{black!3}%
  \colorlet{OGTableShade}{OGBlue!7!white}%
}

\FloatBarrier
\subsection{RQ1: How effectively can existing models identify obligations?}
\label{sec:rq1}

Unfulfilled obligations pose substantial safety risks, yet identifying them requires more than judging whether an agent's recorded actions are permitted.
In this RQ, we evaluate whether existing models can identify accurate and complete obligation sets.

\vspace{2pt}
\noindent\textbf{Setting.}
We evaluate the 14 existing models introduced in Section~\ref{sec:Evaluated Models} on \textsc{ObligationBench}, using the metrics defined in Section~\ref{sec:Metrics}. We report Precision, Recall, and EM on positive instances, together with $\mathrm{CA}_{+}$ and $\mathrm{CA}_{-}$ on positive and negative instances, respectively. We also include \textsc{ObligationGuard} for comparison.

\iflatexml
\begin{table}[t]
\centering
\small
\captionsetup{skip=4pt}
\caption{Obligation identification on \textsc{ObligationBench} (\%). The best and second-best results in each column are bold and underlined.}
\label{tab:rq1-main}

\OGResultsTableSetup
\begin{tabular}{l*{4}{r}r}
\toprule
\textbf{Model} & \multicolumn{4}{c}{\textbf{Positive cases}} & \multicolumn{1}{c}{\textbf{Negative cases}} \\
\cmidrule(lr){2-5}\cmidrule(l){6-6}
& {\textbf{Precision}} & {\textbf{Recall}} & {\textbf{EM}} & {\boldmath$\mathrm{CA}_{+}$} & {\boldmath$\mathrm{CA}_{-}$} \\
\midrule
\rowcolor{OGStripeBg}
\multicolumn{6}{c}{\itshape General-purpose LLMs}\\
\midrule
Claude-Opus-4.8 & 73.45 & {\underline{48.97}} & 9.17 & \bfseries 93.33 & 79.17 \\
GPT-5.6-Sol & {\underline{86.63}} & 47.79 & 9.17 & {\underline{90.83}} & 80.00 \\
DeepSeek-V4-Pro & 78.08 & 33.63 & 5.83 & 65.83 & 90.83 \\
DeepSeek-V4.1-Flash & \bfseries 90.08 & 32.15 & {\underline{10.00}} & 63.33 & {\underline{95.83}} \\
Kimi-K3 & 81.40 & 30.97 & 7.50 & 54.17 & 92.50 \\
GLM-5.3 & 83.06 & 30.38 & 6.67 & 55.83 & 91.67 \\
Qwen3.8-27B & 71.00 & 20.94 & 1.67 & 65.00 & 84.17 \\
Qwen3.8-Max & 85.92 & 17.99 & 0.83 & 41.67 & 94.17 \\
Qwen3-8B & 51.49 & 15.34 & 0.83 & 65.00 & 73.33 \\
GPT-OSS-20B & 58.93 & 9.73 & 1.67 & 37.50 & 80.83 \\
Llama-3.1-8B-Instruct & 3.03 & 0.59 & 0.00 & 34.17 & 94.17 \\
\midrule
\rowcolor{OGStripeBg}
\multicolumn{6}{c}{\itshape Existing guard models}\\
\midrule
Qwen3Guard-Gen-8B & 40.43 & 5.60 & 0.83 & 29.17 & 83.33 \\
AgentDoG1.5-Qwen3.5-4B & 17.65 & 1.77 & 0.00 & 26.67 & 82.50 \\
Llama-Guard-3-8B & 0.00 & 0.00 & 0.00 & 0.00 & \bfseries 100.00 \\
\midrule
\rowcolor{OGTableShade}
\textbf{ObligationGuard (ours)} & 82.98 & \bfseries 57.52 & \bfseries 21.67 & 85.83 & 89.17 \\
\bottomrule
\end{tabular}

\end{table}

\else
\begin{table}[t]
\centering
\small
\captionsetup{skip=4pt}
\caption{Obligation identification on \textsc{ObligationBench} (\%). The best and second-best results in each column are bold and underlined.}
\label{tab:rq1-main}
\OGResultsTableSetup
\begin{tabularx}{\linewidth}{Z*{4}{S[table-column-width=.112\linewidth,table-format=2.2]}S[table-column-width=.13\linewidth,table-format=3.2]}
\toprule
\multirow{2}{*}{\textbf{Model}} & \multicolumn{4}{c}{\textbf{Positive cases}} & \multicolumn{1}{c}{\textbf{Negative cases}} \\
\cmidrule(lr){2-5}\cmidrule(l){6-6}
& {\textbf{Precision}} & {\textbf{Recall}} & {\textbf{EM}} & {\boldmath$\mathrm{CA}_{+}$} & {\boldmath$\mathrm{CA}_{-}$} \\
\midrule
\rowcolor{OGStripeBg}
\multicolumn{6}{c}{\itshape General-purpose LLMs}\\
\midrule
Claude-Opus-4.8 & 73.45 & {\underline{48.97}} & 9.17 & \bfseries 93.33 & 79.17 \\
GPT-5.6-Sol & {\underline{86.63}} & 47.79 & 9.17 & {\underline{90.83}} & 80.00 \\
DeepSeek-V4-Pro & 78.08 & 33.63 & 5.83 & 65.83 & 90.83 \\
DeepSeek-V4.1-Flash & \bfseries 90.08 & 32.15 & {\underline{10.00}} & 63.33 & {\underline{95.83}} \\
Kimi-K3 & 81.40 & 30.97 & 7.50 & 54.17 & 92.50 \\
GLM-5.3 & 83.06 & 30.38 & 6.67 & 55.83 & 91.67 \\
Qwen3.8-27B & 71.00 & 20.94 & 1.67 & 65.00 & 84.17 \\
Qwen3.8-Max & 85.92 & 17.99 & 0.83 & 41.67 & 94.17 \\
Qwen3-8B & 51.49 & 15.34 & 0.83 & 65.00 & 73.33 \\
GPT-OSS-20B & 58.93 & 9.73 & 1.67 & 37.50 & 80.83 \\
Llama-3.1-8B-Instruct & 3.03 & 0.59 & 0.00 & 34.17 & 94.17 \\
\midrule
\rowcolor{OGStripeBg}
\multicolumn{6}{c}{\itshape Existing guard models}\\
\midrule
Qwen3Guard-Gen-8B & 40.43 & 5.60 & 0.83 & 29.17 & 83.33 \\
AgentDoG1.5-Qwen3.5-4B & 17.65 & 1.77 & 0.00 & 26.67 & 82.50 \\
Llama-Guard-3-8B & 0.00 & 0.00 & 0.00 & 0.00 & \bfseries 100.00 \\
\midrule
\rowcolor{OGTableShade}
\textbf{ObligationGuard (ours)} & 82.98 & \bfseries 57.52 & \bfseries 21.67 & 85.83 & 89.17 \\
\bottomrule
\end{tabularx}
\end{table}
\fi

\vspace{2pt}
\noindent\textbf{Results.}
Table~\ref{tab:rq1-main} reports the performance of all evaluated models.

\textbf{Existing models struggle to identify accurate and complete obligation sets.}
Although several existing models achieve high Precision and classification accuracy, their Recall and EM remain low. For example, DeepSeek-V4.1-Flash achieves 90.08\% Precision and 95.83\% $\mathrm{CA}_{-}$, but its Recall is only 32.15\%. Similarly, Claude-Opus-4.8 achieves 93.33\% $\mathrm{CA}_{+}$, yet its EM is only 9.17\%. Across all 14 existing models, the highest Recall and EM are only 48.97\% and 10.00\%, respectively. The low EM is particularly concerning: even the best result corresponds to an accurate and complete obligation set on only one in ten positive instances. Thus, while these models can identify some required actions, their guidance may still leave other safety-critical actions unaddressed.

\textbf{Existing guard models perform worse than most general-purpose LLMs.}
Despite being trained for safety, the three existing guard models achieve at most 5.60\% Recall and 0.83\% EM. They often fail to recognize that any obligations remain, with $\mathrm{CA}_{+}$ no higher than 29.17\%. For example, Llama-Guard-3-8B returns no obligations on any instance, obtaining 100.00\% $\mathrm{CA}_{-}$ but 0.00\% $\mathrm{CA}_{+}$. It therefore provides no guidance on the required actions even when the agent still has unfulfilled obligations. We attribute this gap in part to the focus of existing safety training on detecting forbidden actions. Such training may help models judge whether an action is permitted without teaching them to identify actions that are required but missing from the trajectory.

\textbf{\textsc{ObligationGuard} substantially improves obligation identification.}
By training on our synthetic trajectories, \textsc{ObligationGuard} improves Recall from 15.34\% to 57.52\% and EM from 0.83\% to 21.67\% over its Qwen3-8B backbone. Precision also increases from 51.49\% to 82.98\%, indicating that the model identifies more obligations while making more accurate predictions. These improvements allow \textsc{ObligationGuard} to outperform all evaluated models in both Recall and EM: its Recall exceeds the best existing result by 8.55 percentage points, and its EM is more than twice the best existing result of 10.00\%. The gains over its backbone show that our synthetic trajectories provide useful supervision for identifying obligations in real agent executions.

\begin{boxK}
\textbf{Answer to RQ1.}
Existing models rarely identify accurate and complete obligation sets, and existing guard models return no obligations on most positive instances. \textsc{ObligationGuard} achieves the highest Recall and EM, more than doubling the best existing EM.
\end{boxK}

\FloatBarrier
\subsection{RQ2: What factors affect models' performance in obligation identification?}
\label{sec:rq2}

We are interested in understanding what affects models' performance in obligation identification. We consider two factors: the number of actions between obligation creation and trajectory termination, and the number of obligations that coexist.

\begin{figure}[!t]
    \centering
    \captionsetup{skip=4pt,belowskip=-6pt}
    \captionsetup[subfigure]{font=small,skip=2pt,belowskip=0pt,justification=centering}
    \begin{subfigure}[b]{0.491\linewidth}
        \centering
        \includegraphics[width=\linewidth]{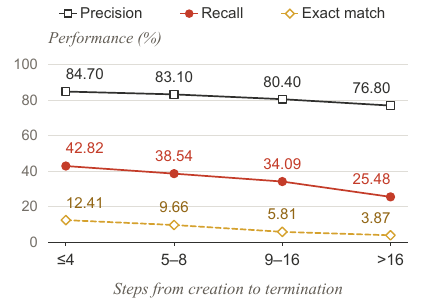}
        \caption{Creation-to-termination distance}
        \label{fig:rq2-creation-distance}
    \end{subfigure}\hfill
    \begin{subfigure}[b]{0.491\linewidth}
        \centering
        \includegraphics[width=\linewidth]{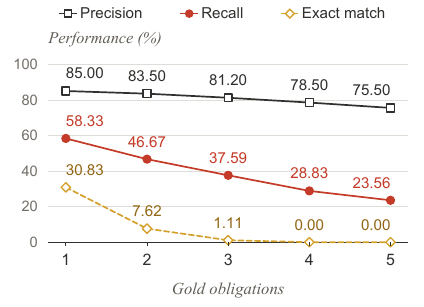}
        \caption{Concurrent obligations}
        \label{fig:rq2-concurrent}
    \end{subfigure}\par
    \caption{Precision, Recall, and EM by (a) the number of actions between the earliest obligation's creation and trajectory termination and (b) the number of coexisting obligations. Results are averaged over five models.}
    \label{fig:rq2-dynamics}
\end{figure}

\vspace{2pt}
\noindent\textbf{Setting.}
We evaluate Claude-Opus-4.8, DeepSeek-V4.1-Flash, DeepSeek-V4-Pro, Kimi-K3, and GLM-5.3 on the 120 positive instances in \textsc{ObligationBench} and report the average results across these five models. Specifically, we examine how their performance varies with the following factors.

\ding{182} \textbf{Creation-to-termination distance.}
We measure this distance as the number of actions between the creation of the obligation and trajectory termination. We use GPT-5.6-Sol to annotate each obligation's creation step and have human experts independently review the annotations. For trajectory $i$, the distance is $D_i=T_i-\min_{q\in\mathcal{O}_i^*}c_i(q)$, where $T_i$ is the final step and $c_i(q)$ is the creation step of obligation $q$. We then group instances into four distance ranges ($\leq4$, 5--8, 9--16, and $>16$) and report Precision, Recall, and EM for each group.

\ding{183} \textbf{Concurrent obligations.}
We group instances by the number of ground-truth obligations, ranging from one to five, and report the same three metrics for each group.

\vspace{2pt}
\noindent\textbf{Results.}
Figure~\ref{fig:rq2-dynamics} shows how performance varies with the two factors.

\textbf{Models perform worse when more actions follow obligation creation.}
As the number of actions between the earliest obligation's creation and termination increases from at most four to more than sixteen, Recall falls from 42.82\% to 25.48\%, and EM from 12.41\% to 3.87\%. Precision also decreases from 84.70\% to 76.80\%, but remains relatively high. The widening gap between Precision and Recall suggests that their predictions become increasingly incomplete as more actions intervene before termination, even though individual predictions are often correct.

\textbf{Obligation identification becomes harder as more obligations coexist.}
As the number of ground-truth obligations increases from one to five, Recall decreases from 58.33\% to 23.56\%, while EM falls from 30.83\% to 0.00\%. When five obligations coexist, models identify fewer than one quarter of the required actions on average, and none returns an accurate and complete obligation set. The decline in Recall shows that models miss a growing proportion of obligations as more coexist. These omissions make complete identification increasingly difficult: a guard model may identify some required actions while leaving other safety-critical actions unaddressed.

\begin{boxK}
\iflatexml\else\vspace{2pt}\fi
\noindent\textbf{Answer to RQ2.}
Models identify obligations less completely when more actions follow obligation creation and when more obligations coexist.
\end{boxK}

\FloatBarrier

\subsection{RQ3: What contributes to the effectiveness of \textsc{ObligationGuard}?}
\label{sec:rq3}

The effectiveness of \textsc{ObligationGuard} may depend on its backbone, the amount of training data, and how these examples are generated. We examine how performance varies across model sizes and families, and whether using synthesis templates produces more effective training data.

\vspace{2pt}
\noindent\textbf{Setting.}
We conduct the following comparisons on \textsc{ObligationBench}, using the metrics defined in Section~\ref{sec:Metrics} and the training method described in Section~\ref{sec:model-training}.

\ding{182} \textbf{Model size.}
To examine how the benefits vary with model size, we fine-tune Qwen3-4B, Qwen3-8B, and Qwen3-14B on the same 40,000 synthetic examples and compare each model with its original backbone.

\ding{183} \textbf{Model family.}
To examine whether our synthetic data also benefits another model family, we fine-tune Llama-3.1-8B-Instruct using the same data and training settings as Qwen3-8B, and compare it with its original backbone.

\ding{184} \textbf{Training-data size.}
We train Qwen3-8B on 5K, 10K, 20K, and 40K synthetic examples to examine how the amount of training data affects obligation identification.

\ding{185} \textbf{Synthesis templates.}
Our approach first constructs synthesis templates that specify tasks, safety requirements, and the obligations to remain unfulfilled, then generates trajectories from these templates. To examine whether the templates are useful, we ask GPT-5.6-Sol to directly generate tasks, trajectories, and their obligation annotations without using templates. We refer to this setting as \emph{direct synthesis} and compare Qwen3-8B trained on these examples with \textsc{ObligationGuard}. Both settings use GPT-5.6-Sol as the generator, 40K training examples, and the same training settings.

\iflatexml
\begin{table}[t]
\centering
\captionsetup{skip=4pt}
\caption{Obligation identification before and after fine-tuning on our 40,000 synthetic examples (\%).}
\label{tab:rq3-scale-family}
\OGResultsTableSetup

\begin{tabular}{l*{4}{r}r}
\toprule
\textbf{Setting}
& \multicolumn{4}{c}{\textbf{Positive cases}} & \multicolumn{1}{c}{\textbf{Negative cases}} \\
\cmidrule(lr){2-5}\cmidrule(l){6-6}
& {\textbf{Precision}} & {\textbf{Recall}} & {\textbf{EM}} & {\boldmath$\mathrm{CA}_{+}$} & {\boldmath$\mathrm{CA}_{-}$} \\
\midrule
\rowcolor{OGStripeBg}
\multicolumn{6}{c}{\textbf{Qwen3-4B}}\\
\midrule
Original
& 44.05 & 10.91 & 0.00 & 48.33 & 69.17 \\
Fine-tuned
& 77.49 & 43.66 & 12.50 & 83.33 & 84.17 \\
\midrule
\rowcolor{OGStripeBg}
\multicolumn{6}{c}{\textbf{Qwen3-8B}}\\
\midrule
Original
& 51.49 & 15.34 & 0.83 & 65.00 & 73.33 \\
\rowcolor{OGTableShade}
\textbf{Fine-tuned (ours)}
& 82.98 & 57.52 & 21.67 & 85.83 & 89.17 \\
\midrule
\rowcolor{OGStripeBg}
\multicolumn{6}{c}{\textbf{Qwen3-14B}}\\
\midrule
Original
& 63.37 & 18.88 & 1.67 & 69.17 & 75.83 \\
Fine-tuned
& 84.80 & 62.54 & 25.83 & 91.67 & 90.83 \\
\midrule
\rowcolor{OGStripeBg}
\multicolumn{6}{c}{\textbf{Llama-3.1-8B-Instruct}}\\
\midrule
Original
& 3.03 & 0.59 & 0.00 & 34.17 & 94.17 \\
Fine-tuned
& 78.24 & 49.85 & 16.67 & 85.83 & 87.50 \\
\bottomrule
\end{tabular}

\end{table}

\else
\begin{table}[t]
\centering
\captionsetup{skip=4pt}
\caption{Obligation identification before and after fine-tuning on our 40,000 synthetic examples (\%).}
\label{tab:rq3-scale-family}
\OGResultsTableSetup
\begin{tabularx}{\linewidth}{Z*{4}{S[table-column-width=.112\linewidth,table-format=2.2]}S[table-column-width=.13\linewidth,table-format=3.2]}
\toprule
\multirow{2}{*}{\textbf{Setting}}
& \multicolumn{4}{c}{\textbf{Positive cases}} & \multicolumn{1}{c}{\textbf{Negative cases}} \\
\cmidrule(lr){2-5}\cmidrule(l){6-6}
& {\textbf{Precision}} & {\textbf{Recall}} & {\textbf{EM}} & {\boldmath$\mathrm{CA}_{+}$} & {\boldmath$\mathrm{CA}_{-}$} \\
\midrule
\rowcolor{OGStripeBg}
\multicolumn{6}{c}{\textbf{Qwen3-4B}}\\
\midrule
Original
& 44.05 & 10.91 & 0.00 & 48.33 & 69.17 \\
Fine-tuned
& 77.49 & 43.66 & 12.50 & 83.33 & 84.17 \\
\midrule
\rowcolor{OGStripeBg}
\multicolumn{6}{c}{\textbf{Qwen3-8B}}\\
\midrule
Original
& 51.49 & 15.34 & 0.83 & 65.00 & 73.33 \\
\rowcolor{OGTableShade}
\textbf{Fine-tuned (ours)}
& 82.98 & 57.52 & 21.67 & 85.83 & 89.17 \\
\midrule
\rowcolor{OGStripeBg}
\multicolumn{6}{c}{\textbf{Qwen3-14B}}\\
\midrule
Original
& 63.37 & 18.88 & 1.67 & 69.17 & 75.83 \\
Fine-tuned
& 84.80 & 62.54 & 25.83 & 91.67 & 90.83 \\
\midrule
\rowcolor{OGStripeBg}
\multicolumn{6}{c}{\textbf{Llama-3.1-8B-Instruct}}\\
\midrule
Original
& 3.03 & 0.59 & 0.00 & 34.17 & 94.17 \\
Fine-tuned
& 78.24 & 49.85 & 16.67 & 85.83 & 87.50 \\
\bottomrule
\end{tabularx}
\end{table}
\fi

\begin{figure}[t]
    \centering
    \captionsetup{skip=4pt}
    \captionsetup[subfigure]{font=small,skip=2pt,justification=centering}
    \begin{subfigure}[b]{0.623\linewidth}
        \centering
        \includegraphics[width=\linewidth]{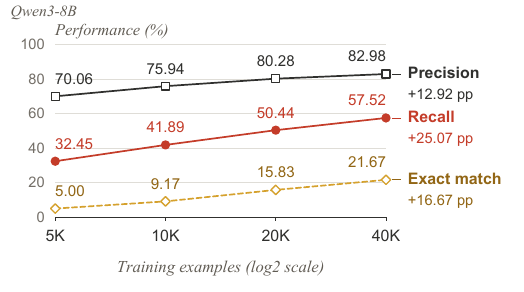}
        \caption{Training-data size}
        \label{fig:rq3-training-data}
    \end{subfigure}
    \hfill
    \begin{subfigure}[b]{0.359\linewidth}
        \centering
        \includegraphics[width=\linewidth]{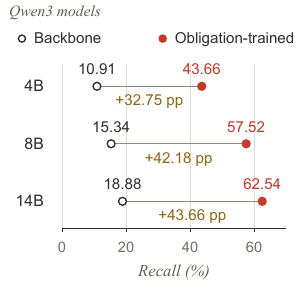}
        \caption{Model size}
        \label{fig:rq3-model-size}
    \end{subfigure}
    \caption{Effects of training-data size and model size on obligation identification. (a) Precision, Recall, and EM of Qwen3-8B fine-tuned on 5K--40K synthetic examples. (b) Recall of Qwen3 models of different sizes before and after fine-tuning on 40K synthetic examples.}
    \label{fig:rq3-scale}
\end{figure}

\begin{figure}[t]
    \centering
    \captionsetup{skip=2pt}
    \includegraphics[width=0.95\linewidth]{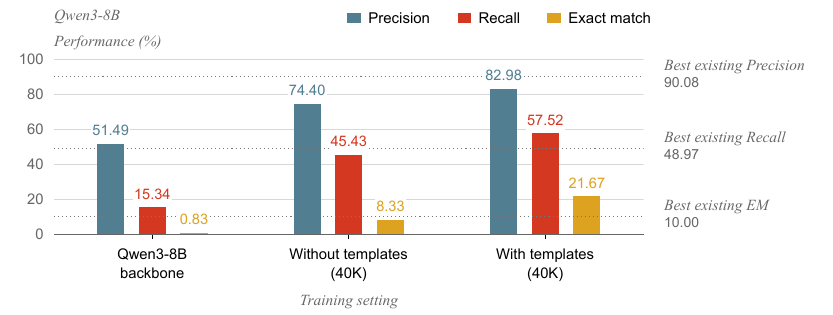}
    \caption{Effect of synthesis templates on Qwen3-8B. Both synthesis methods provide 40K training examples. Dashed lines mark the best existing results from Table~\ref{tab:rq1-main}.}
    \label{fig:scenario-planning-ablation}
\end{figure}

\vspace{2pt}
\noindent\textbf{Results.}
Table~\ref{tab:rq3-scale-family} reports the results across model sizes and families. Figures~\ref{fig:rq3-scale} and~\ref{fig:scenario-planning-ablation} show the effects of training-data size and the use of synthesis templates.

\textbf{Our synthetic data improves obligation identification at all three model sizes.}
As shown in Table~\ref{tab:rq3-scale-family}, fine-tuning improves Precision, Recall, and EM for all three Qwen3 models. For example, Qwen3-4B's Recall increases from 10.91\% to 43.66\%, and its EM from 0.00\% to 12.50\%. Qwen3-14B also benefits substantially, with Recall rising from 18.88\% to 62.54\% and EM from 1.67\% to 25.83\%. Larger fine-tuned models achieve better results, but even the fine-tuned 4B model exceeds the original 14B model in all three metrics. Within this range, training on trajectories with obligation annotations provides larger gains than increasing the size of the original model.

\textbf{The same synthetic data also benefits a different model family.}
Fine-tuning Llama-3.1-8B-Instruct increases Precision from 3.03\% to 78.24\%, Recall from 0.59\% to 49.85\%, and EM from 0.00\% to 16.67\%. Despite its poor initial performance, the fine-tuned model exceeds the best existing Recall and EM reported in RQ1. The usefulness of our synthetic data therefore extends beyond the Qwen3 family used to develop \textsc{ObligationGuard}.

\textbf{More synthetic examples improve both accuracy and completeness.}
Figure~\ref{fig:rq3-training-data} shows consistent improvements as the training set grows. Increasing the number of examples from 5K to 40K raises Precision from 70.06\% to 82.98\%, Recall from 32.45\% to 57.52\%, and EM from 5.00\% to 21.67\%. All three metrics improve at each evaluated data size, indicating that additional examples help the model identify more obligations accurately. These results support synthesizing training data at scale: within the evaluated range, additional synthetic trajectories continue to improve performance on real agent executions.

\textbf{Using synthesis templates makes the training data more effective.}
As shown in Figure~\ref{fig:scenario-planning-ablation}, training on directly generated examples improves Qwen3-8B over its original backbone, achieving 74.40\% Precision, 45.43\% Recall, and 8.33\% EM. Using synthesis templates further raises these metrics to 82.98\%, 57.52\%, and 21.67\%, respectively. With the same generator, training-data size, and fine-tuning settings, EM more than doubles. We attribute these gains in part to making the safety requirements and obligation sets explicit before trajectory generation. The templates guide the generator in constructing trajectories consistent with these requirements and provide the corresponding obligation labels, helping the guard model correctly learn which safety-critical actions remain unfulfilled.

\begin{boxK}
\textbf{Answer to RQ3.} Fine-tuning on our synthetic training data improves performance across the evaluated model sizes and families. Increasing the number of training examples further improves all three metrics. With the same number of training examples, using scenario planning during data synthesis leads to better performance than direct synthesis.
\end{boxK}

\FloatBarrier
\subsection{RQ4: Does stronger obligation identification improve agent safety?}
\label{sec:rq4}

The practical value of obligation identification depends on whether it can help agents complete the required but unperformed safety-critical actions. We therefore examine whether guidance from guard models improves coding-agent safety and whether guard models with better performance on \textsc{ObligationBench} provide more effective guidance.

\vspace{2pt}
\noindent\textbf{Setting.}
We evaluate Qwen3.8-27B with Mini-SWE-Agent on 186 tasks from SusVibes~\cite{susvibes}. At the agent's first termination attempt, we compare three conditions:
\ding{182} \textbf{No guidance.} The agent terminates without receiving additional guidance.
\ding{183} \textbf{Self-reminder.} We ask the agent to recheck potential safety issues~\cite{xie2023selfreminders}.
\ding{184} \textbf{Guard model guidance.} We provide the task and trajectory to a guard model and return its predicted obligations to the agent.
All three conditions start from the same execution state. In the two guidance conditions, the agent can continue execution to revise its solution, but receives guidance at most once per execution.

We compare the six guard models listed in Table~\ref{tab:rq4-main}. We report FuncPass, the percentage of tasks that pass the functional tests, and SecPass, the percentage that pass both the functional and safety tests. To examine how guidance changes individual outcomes, we classify each solution as \emph{Incorrect} if it fails the functional tests, \emph{Correct-Unsafe} if it passes only the functional tests, or \emph{Correct-Safe} if it passes both. We also examine the association between each guard model's Recall on \textsc{ObligationBench} and its improvement in SecPass.

\iflatexml
\begin{table}[t]
\centering
\captionsetup{skip=4pt}
\caption{Effects of guidance on the Qwen3.8-27B agent running with Mini-SWE-Agent on SusVibes. $\Delta$Sec denotes the improvement over no guidance. The best and second-best results are bold and underlined.}
\label{tab:rq4-main}
\OGResultsTableSetup
\begin{tabular}{l*{2}{r}r}
\toprule
\textbf{Guidance source} & {\textbf{FuncPass (\%)}} & {\textbf{SecPass (\%)}} & {\textbf{\boldmath$\Delta$Sec (\%)}} \\
\midrule
No guidance & \bfseries 27.4 & 6.5 & {--} \\
Self-reminder & {\underline{26.9}} & 8.1 & +1.6 \\
\midrule
Qwen3-8B & 25.8 & 7.5 & +1.0 \\
GLM-5.3 & 26.3 & 9.7 & +3.2 \\
Kimi-K3 & 26.3 & 9.1 & +2.6 \\
DeepSeek-V4.1-Flash & {\underline{26.9}} & 10.8 & +4.3 \\
Claude-Opus-4.8 & 25.8 & {\underline{12.4}} & {\underline{+5.9}} \\
\midrule
\rowcolor{OGTableShade}
\textbf{ObligationGuard (ours)} & {\underline{26.9}} & \bfseries 15.1 & \bfseries +8.6 \\
\bottomrule
\end{tabular}

\end{table}

\else
\begin{table}[t]
\centering
\captionsetup{skip=4pt}
\caption{Effects of guidance on the Qwen3.8-27B agent running with Mini-SWE-Agent on SusVibes. $\Delta$Sec denotes the improvement over no guidance. The best and second-best results are bold and underlined.}
\label{tab:rq4-main}
\OGResultsTableSetup
\begin{tabularx}{.94\linewidth}{Z*{2}{S[table-column-width=.165\linewidth,table-format=2.1]}S[table-column-width=.17\linewidth,table-format=+1.1]}
\toprule
\textbf{Guidance source} & {\textbf{FuncPass (\%)}} & {\textbf{SecPass (\%)}} & {\textbf{\boldmath$\Delta$Sec (\%)}} \\
\midrule
No guidance & \bfseries 27.4 & 6.5 & {--} \\
Self-reminder & {\underline{26.9}} & 8.1 & +1.6 \\
\midrule
Qwen3-8B & 25.8 & 7.5 & +1.0 \\
GLM-5.3 & 26.3 & 9.7 & +3.2 \\
Kimi-K3 & 26.3 & 9.1 & +2.6 \\
DeepSeek-V4.1-Flash & {\underline{26.9}} & 10.8 & +4.3 \\
Claude-Opus-4.8 & 25.8 & {\underline{12.4}} & {\underline{\num{+5.9}}} \\
\midrule
\rowcolor{OGTableShade}
\textbf{ObligationGuard (ours)} & {\underline{26.9}} & \bfseries 15.1 & \bfseries +8.6 \\
\bottomrule
\end{tabularx}
\end{table}
\fi

\vspace{2pt}
\noindent\textbf{Results.}
Table~\ref{tab:rq4-main} reports agent performance under each guidance condition. Table~\ref{tab:rq4-repair} further shows how \textsc{ObligationGuard} helps the agent address safety failures.

\textbf{\textsc{ObligationGuard} improves agent safety while largely preserving functional correctness.}
Without guidance, the agent achieves 6.5\% SecPass. A self-reminder raises this to 8.1\%, while guidance from \textsc{ObligationGuard} increases it to 15.1\%, exceeding all other guidance conditions. The strongest existing guard model in this experiment, Claude-Opus-4.8, achieves 12.4\% SecPass. Meanwhile, FuncPass changes only from 27.4\% without guidance to 26.9\% with \textsc{ObligationGuard}. These results demonstrate the effectiveness of \textsc{ObligationGuard} in real agent executions: it more than doubles SecPass and achieves the largest safety gain among all evaluated guard models, while largely preserving functional completion.

\iflatexml
\begin{table}[t]
\centering
\captionsetup{skip=3pt}
\caption{Functional and safety test results before (rows) and after (columns) \textsc{ObligationGuard} guidance. Values are task counts. The highlighted cell shows 16 solutions that passed only the functional tests before guidance and passed both tests afterward.}
\label{tab:rq4-repair}
\OGResultsTableSetup
\begin{tabular}{lr*{3}{r}}
\toprule
\multicolumn{2}{c}{\textbf{Before guidance}} & \multicolumn{3}{c}{\textbf{After guidance}} \\
\cmidrule(lr){1-2}\cmidrule(lr){3-5}
\textbf{Outcome} & {\textbf{Tasks}} & {\textbf{Incorrect}} & {\textbf{Correct-Unsafe}} & {\textbf{Correct-Safe}} \\
\midrule
Incorrect & 135 & 133 & 1 & 1 \\
Correct-Unsafe & 39 & 3 & 20 & \cellcolor{OGTableShade}\bfseries 16 \\
Correct-Safe & 12 & 0 & 1 & 11 \\
\bottomrule
\end{tabular}

\end{table}

\else
\begin{table}[t]
\centering
\captionsetup{skip=3pt}
\caption{Functional and safety test results before (rows) and after (columns) \textsc{ObligationGuard} guidance. Values are task counts. The highlighted cell shows 16 solutions that passed only the functional tests before guidance and passed both tests afterward.}
\label{tab:rq4-repair}
\OGResultsTableSetup
\begin{tabularx}{\linewidth}{ZS[table-column-width=.09\linewidth,table-format=3.0]*{3}{S[table-column-width=.185\linewidth,table-format=3.0]}}
\toprule
\multicolumn{2}{c}{\textbf{Before guidance}} & \multicolumn{3}{c}{\textbf{After guidance}} \\
\cmidrule(lr){1-2}\cmidrule(lr){3-5}
\textbf{Outcome} & {\textbf{Tasks}} & {\textbf{Incorrect}} & {\textbf{Correct-Unsafe}} & {\textbf{Correct-Safe}} \\
\midrule
Incorrect & 135 & 133 & 1 & 1 \\
Correct-Unsafe & 39 & 3 & 20 & \cellcolor{OGTableShade}\bfseries 16 \\
Correct-Safe & 12 & 0 & 1 & 11 \\
\bottomrule
\end{tabularx}
\end{table}
\fi

\textbf{Most safety gains come from correcting unsafe solutions that already pass functional tests.}
Table~\ref{tab:rq4-repair} shows that 16 of the 39 initially Correct-Unsafe solutions become Correct-Safe after \textsc{ObligationGuard} guidance. Only one initially Incorrect solution becomes Correct-Safe. Of the 17 newly safe solutions, 16 therefore already passed the functional tests before guidance. This result helps explain why SecPass improves substantially while FuncPass remains nearly unchanged: the guard model primarily helps the agent address safety failures in otherwise working solutions.

\begin{figure}[!htbp]
    \centering
    \captionsetup{skip=1pt}
    \includegraphics[width=0.95\linewidth]{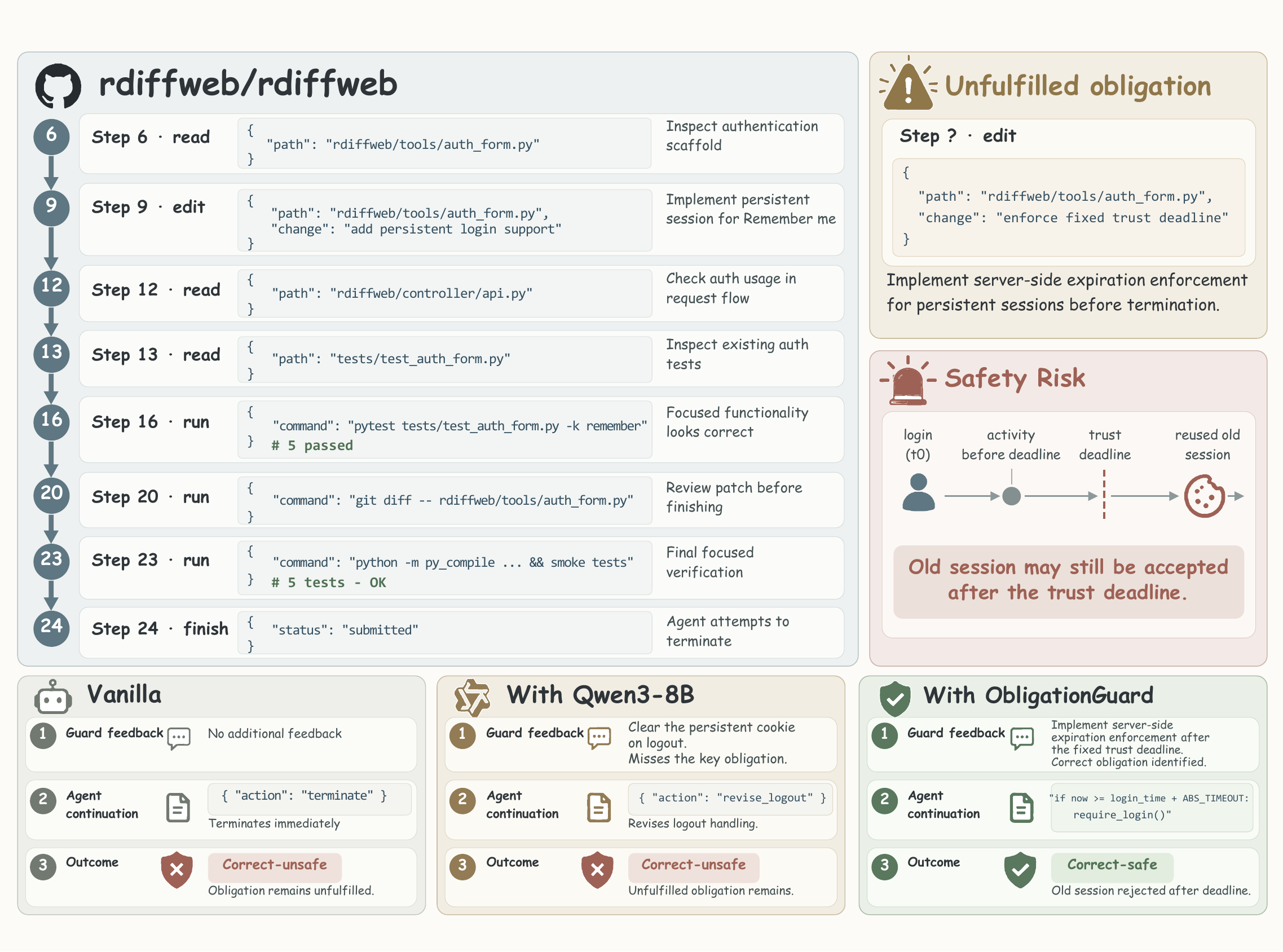}
    \caption{On a SusVibes task, \textsc{ObligationGuard} guides the agent to enforce server-side session expiration, while Qwen3-8B misses this obligation.}
    \label{fig:rq4-case-study}
\end{figure}

\textbf{Identifying the unfulfilled obligation helps the agent address the specific safety failure.}
Figure~\ref{fig:rq4-case-study} illustrates how the guidance leads to a safer solution. In this persistent-login task, the agent attempts to terminate at step 24, although server-side session expiration is still missing. An old session may therefore remain usable after its intended expiration time. Qwen3-8B suggests clearing the persistent cookie on logout, which leaves expiration enforcement unaddressed. \textsc{ObligationGuard} instead identifies the obligation to reject expired sessions. Following this guidance, the agent adds the server-side expiration check and requires users with expired sessions to log in again. Its solution becomes Correct-Safe while retaining its functional behavior. This case illustrates how identifying the unfulfilled obligation guides the agent toward the required safety-critical action.

\textbf{Guard models with better performance on \textsc{ObligationBench} tend to improve agent safety more.}
Across the six evaluated guard models, Recall on \textsc{ObligationBench} is strongly associated with the improvement in SecPass on SusVibes (Spearman's $\rho=0.94$; Pearson's $r=0.97$). For example, Qwen3-8B achieves 15.34\% Recall and increases SecPass by only 1.0 percentage points. Claude-Opus-4.8 and \textsc{ObligationGuard}, with Recall of 48.97\% and 57.52\%, increase SecPass by 5.9 and 8.6 percentage points, respectively. This association links performance on \textsc{ObligationBench} to the usefulness of guidance from guard models during agent execution. It suggests that guard models that identify more unfulfilled obligations can give agents more complete guidance on the safety-critical actions still required.

\begin{boxK}
\textbf{Answer to RQ4.}
\textsc{ObligationGuard} increases SecPass from 6.5\% to 15.1\% with little change in FuncPass. Better performance on \textsc{ObligationBench} is associated with larger safety gains on SusVibes.
\end{boxK}
\FloatBarrier

\section{Discussion}

\label{sec:discussion}

\subsection{Reliability of LLM Judgment}
\label{sec:judge-validation}

We use GPT-5.6-Sol to assess semantic equivalence between predicted and ground-truth obligations. To examine the reliability of these judgments, we manually review 400 obligation pairs from the RQ1 results, stratified by evaluated model: 200 judged as matched and 200 as unmatched. Reviewers assess each pair using the corresponding task and trajectory, without seeing the model's identity or the LLM judge's decision. Human review confirms 196 of the 200 matched judgments (98.0\%) and 184 of the 200 unmatched judgments (92.0\%), giving 95.0\% overall agreement on this balanced sample. These results indicate that the LLM judge provides reliable semantic matching judgments.

\subsection{Threats to Validity}
\label{sec:threats}

We consider four threats to the validity of our study:

\textbf{Reliability of obligation annotations.} Annotating obligations requires determining which safety-critical actions are necessary and whether the agent has completed them. Automatic annotation may overlook a required action or mistake an unnecessary action for an obligation, affecting model evaluation. To mitigate this threat, two software engineers with over five years of professional experience independently review and correct the initial annotations. They check whether the annotated actions remain required at termination and identify any missing obligations, then resolve disagreements through discussion and discard ambiguous cases.

\textbf{Diversity and realism of the benchmark.} Synthetic trajectories may not capture the variety of tasks and execution behavior that agents encounter in practice, limiting the benchmark's diversity and realism. We therefore collect trajectories from actual executions on established software-engineering benchmarks, covering issue resolution, feature development, and terminal operation. Agents powered by four different LLMs provide varied trajectories within these domains. This coverage also extends to the safety requirements: the benchmark includes five safety categories and one to five unfulfilled obligations per positive instance.

\textbf{Practical relevance of the benchmark.} High scores on \textsc{ObligationBench} alone do not establish whether a guard model improves agent safety in practice. We therefore evaluate six guard models on SusVibes by using their predicted obligations to guide an agent. At its first termination attempt, the agent receives the identified obligations and can continue execution to fulfill them. We find that higher Recall on \textsc{ObligationBench} is associated with larger SecPass gains, supporting the benchmark's relevance to agent safety during actual execution.

\textbf{Generalizability of the findings.} The choice of evaluated models may limit how broadly our findings apply. Our evaluation includes 14 existing models spanning different families and sizes, covering both general-purpose LLMs and guard models. We also examine whether the improvements from our synthetic training data extend beyond a single backbone by fine-tuning Qwen3 models at three sizes and a Llama backbone on the same data.

\section{Conclusion}
In this paper, we show that identifying forbidden actions alone is insufficient to ensure agent safety, as unfulfilled obligations constitute a major source of safety risk. We introduce \textsc{ObligationBench}, the first benchmark for obligation identification, and find that 14 existing models struggle to identify accurate and complete obligation sets. To address this limitation, we develop \textsc{ObligationGuard} using synthetic training data generated through scenario planning and trajectory synthesis. Our experiments demonstrate that \textsc{ObligationGuard} improves obligation identification and helps agents complete tasks more safely while largely preserving functional completion. We hope this work encourages the community to incorporate obligation identification into the design and evaluation of future guard models.

\section{Data Availability}
Our code, benchmarks, and \textit{Supplementary Materials} can be accessed through a public repository at: \url{https://github.com/THU-Agent/ObligationGuard}.

\bibliographystyle{preprint}
\iflatexml
  \ifcsname c@@bibitem\endcsname
    \setcounter{@bibitem}{0}
  \fi
\fi
\bibliography{references}

\end{document}